\documentclass[10pt,twocolumn,letterpaper]{article}

\usepackage{iccv}
\usepackage[linesnumbered,ruled]{algorithm2e}
\usepackage{algpseudocode}
\usepackage{times}
\usepackage{epsfig}
\usepackage{graphicx}
\usepackage{color}
\usepackage{graphics} 
\usepackage{epsfig} 
\usepackage{amsmath} 
\usepackage{amssymb}  
\usepackage{epstopdf}
\usepackage{mathtools}
\usepackage{stfloats}
\usepackage{makecell}
\usepackage{multirow}
\DeclareMathOperator*{\argmax}{argmax}

\newcommand{\PreserveBackslash}[1]{\let\temp=\\#1\let\\=\temp}

\newcolumntype{C}[1]{>{\PreserveBackslash\centering}p{#1}}
\newcolumntype{R}[1]{>{\PreserveBackslash\raggedleft}p{#1}}
\newcolumntype{L}[1]{>{\PreserveBackslash\raggedright}p{#1}}

\usepackage{cite}

\iccvfinalcopy 

\def\iccvPaperID{3120} 

\def \TRK{\mathcal{T}}
\def \VRF{\mathcal{V}}

\def \TK{\mathcal{T}}
\def \VF{\mathcal{V}}

\pdfoutput=1
\begin{document}

\title{Parallel Tracking and Verifying: A Framework for Real-Time and High Accuracy Visual Tracking}

\author{Heng Fan ~~~~~~   Haibin Ling\thanks{Corresponding author.}\\
    \normalsize Meitu HiScene Lab, HiScene Information Technologies, Shanghai, China\\
    \normalsize Department of Computer and Information Sciences, Temple University, Philadelphia, PA USA\\
{\tt\small \{hengfan,hbling\}@temple.edu}
}

\maketitle

%
\begin{abstract}
Being intensively studied, visual tracking has seen great recent advances in either speed (\eg, with correlation filters) or accuracy (\eg, with deep features). Real-time and high accuracy tracking algorithms, however, remain scarce. In this paper we study the problem from a new perspective and present a novel \emph{parallel tracking and verifying} (PTAV) framework, by taking advantage of the ubiquity of multi-thread techniques and borrowing from the success of \emph{parallel tracking and mapping} in visual SLAM. Our PTAV framework typically consists of two components, a tracker $\TRK$ and a verifier $\VRF$, working in parallel on two separate threads. The tracker $\TRK$ aims to provide a super real-time tracking inference and is expected to perform well most of the time; by contrast, the verifier $\VRF$ checks the tracking results and corrects $\TRK$ when needed. The key innovation is that, $\VRF$ does not work on every frame but only upon the requests from $\TRK$; on the other end, $\TRK$ may adjust the tracking according to the feedback from $\VRF$. With such collaboration, PTAV enjoys both the high efficiency provided by $\TRK$ and the strong discriminative power by $\VRF$. In our extensive experiments on popular benchmarks including OTB2013, OTB2015, TC128 and UAV20L, PTAV achieves the best tracking accuracy among all real-time trackers, and in fact performs even better than many deep learning based solutions. Moreover, as a general framework, PTAV is very flexible and has great rooms for improvement and generalization.
\end{abstract}

\section{Introduction}

\begin{figure}[!t]
\centering
\begin{tabular}{@{}C{8.1cm}@{}}
\includegraphics[width=8.05cm]{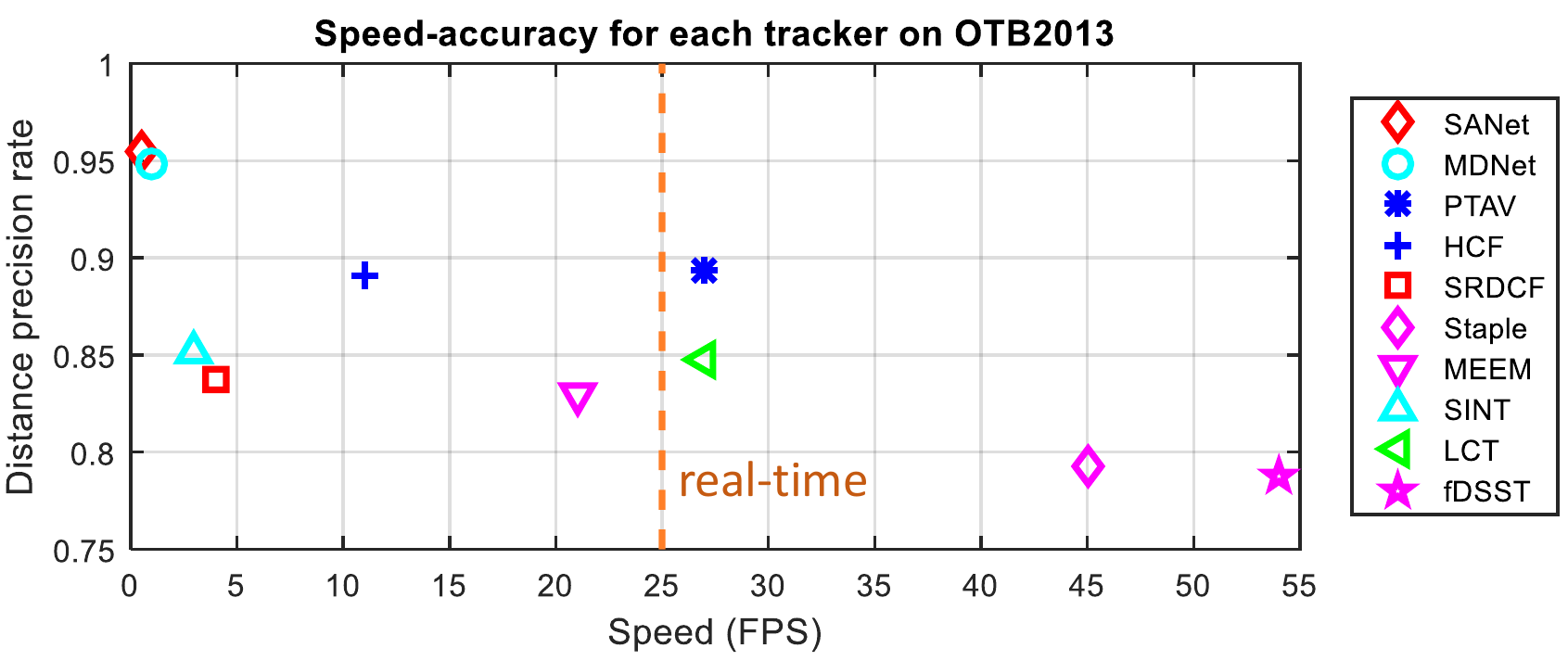}\\
\end{tabular}
\caption{Speed and accuracy plot of state-of-the-art visual trackers on OTB2013~\cite{wu2013online}. For better illustration, only those trackers with accuracy higher than 0.75 are reported. The proposed PTAV algorithm achieves the best accuracy among all real-time trackers.}
\label{fig:speed-accuracy}
\end{figure}

\subsection{Background}

Visual object tracking plays a crucial role in computer vision and has a variety of applications such as robotics, visual surveillance, human-computer interaction and so forth \cite{smeulders2014visual,YilmazJS06survey}. Despite great successes in recent decades, robust visual tracking still remains challenging due to many factors including object deformation, occlusion, rotation, illumination change, pose variation, etc. An emerging trend toward improving tracking accuracy is to use deep learning-based techniques (e.g., \cite{ma2015hierarchical,nam2016learning,fan2016sanet,wang2013learning}) for their strong discriminative power, as shown in~\cite{nam2016learning}. Such algorithms, unfortunately, often suffer from high computational burden and hardly run in real-time (see Fig.~\ref{fig:speed-accuracy}).

Along a somewhat orthogonal direction, researchers have been proposing efficient visual trackers (e.g., \cite{henriques2015high,zhang2012real,danelljan2016discriminative,bolme2010visual}), represented by the series of trackers based on correlation filters. While easily running at real-time, these trackers are usually less robust than deep learning-based approaches.

Despite the above mentioned progresses in either speed or accuracy, real-time high quality tracking algorithms remain scarce. A natural way is to seek a trade-off between speed and accuracy, {\eg,~\cite{bertinetto2016staple,ma2015long}}. In this paper we work toward this goal, but from a novel perspective as following.

\begin{figure*}[!t]
\centering
\begin{tabular}{@{}C{17.1cm}@{}}
\includegraphics[width=16.75cm,height=2cm]{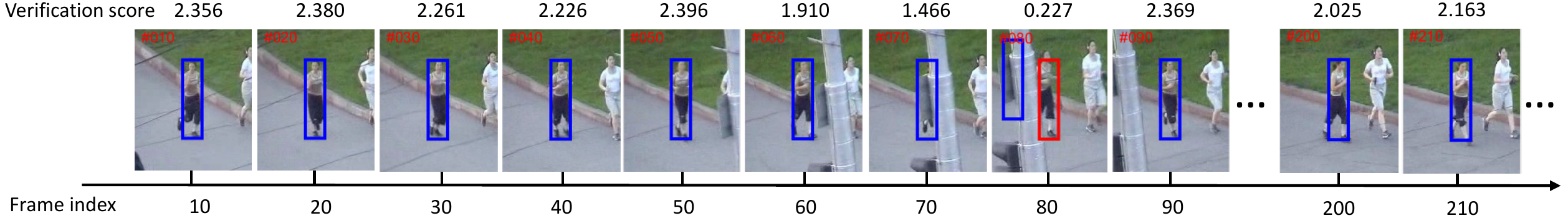}\\
\end{tabular}
\caption{Verifying tracking results on a typical sequence. Verifier validates tracking results every 10 frames. Most of the time the tracking results are reliable (showing in \textcolor{blue}{blue}). Occasionally, \eg frame \#080, the verifier finds the original tracking result (showing in \textcolor{blue}{blue}) unreliable and the tracker is corrected and resumes tracking based on detection result (showing in \textcolor{red}{red}).}
\label{fig:10frames}
\end{figure*}
\subsection{Motivation}

Our key idea is to decompose the original tracking task into two parallel but collaborative ones, one for fast tracking and the other for accurate verification. Our work is inspired by the following observations or related works:
\\
\textbf{Motivation 1.} When tracking a target from visual input, most of the time the target moves smoothly and its appearance changes little or slowly. Simple but efficient algorithms usually work fine for such easy cases. By contrast, hard cases appear only occasionally, though may cause serious consequences if not addressed properly. These hard cases typically require to be handled by computationally expensive operations, which are called verifiers in our study. These verifiers, intuitively, are needed only occasionally instead of for every frame, as shown in Fig.~\ref{fig:10frames}.
\\
\textbf{Motivation 2.} The ubiquity of multi-thread computing has already benefited computer vision systems, with notably in visual SLAM (\emph{simultaneous localization and mapping}). By splitting tracking and mapping into two parallel threads, PTAM (\emph{parallel tracking and mapping})~\cite{klein2007parallel} provides one of the most popular SLAM frameworks with many important extensions~(\eg,~\cite{ORB-SLAM}). A key inspiration in PTAM is that mapping is not needed for every frame; nor does verifying in our task.
\\
\textbf{Motivation 3.} Last but not least, recent advances in fast or accurate tracking algorithms provide promising building blocks and encourage us to seek a combined solution.

\subsection{Contribution}

With the motivations listed above, we propose to build real-time high accuracy trackers in a novel framework named \emph{parallel tracking and verifying} (PTAV). PTAV typically consists of two components: a fast tracker\footnote{In the rest of the paper, for conciseness, we refer the \emph{fast tracker} as a \emph{tracker} whenever no confusion caused.} denoted by $\TRK$ and a verifier denoted by $\VRF$.  The two components work in parallel on two separate threads while collaborate with each other. The tracker $\TRK$ aims to provide a super real-time tracking inference that is expected to perform well most of the time, \eg, most frames in Fig.~\ref{fig:10frames}. By contrast, the verifier $\VRF$ checks the tracking results and corrects $\TRK$ when needed, \eg, at frame \#080 in Fig.~\ref{fig:10frames}.

The key idea is, while $\TRK$ needs to run on every frame, $\VRF$ does not. As a general framework, PTAV allows the coordination between the tracker and the verifier: $\VRF$ checks the tracking results provided by $\TRK$ and sends feedback to $\VRF$; and $\VRF$ adjusts itself according to the feedback when necessary. By running $\TRK$ and $\VRF$ in parallel, PTAV inherits both the high efficiency of $\TRK$ and the strong discriminative power of $\VRF$.

\begin{figure*}[!t]
\centering
\includegraphics[width=.98\linewidth,height=5.5cm]{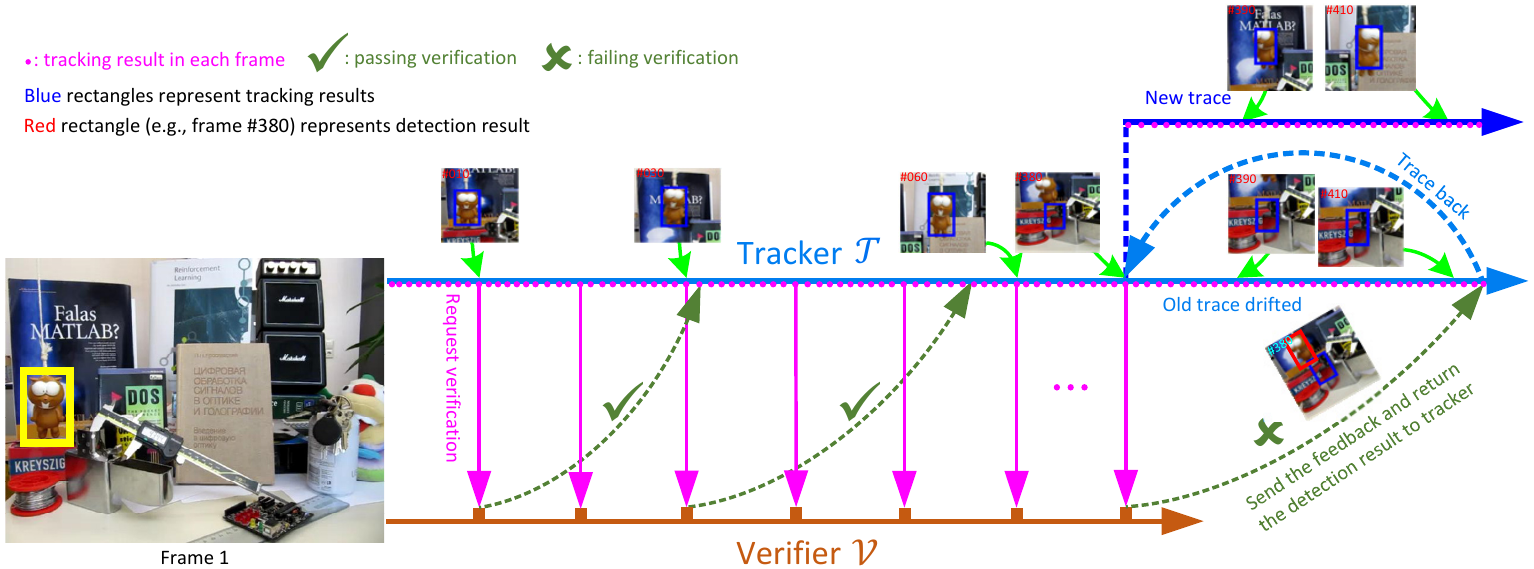}\\
\caption{Illustration of the PTAV framework in which tracking and verifying are processed in two parallel asynchronous threads.}
\label{detail_PTV}
\end{figure*}

Implementing\footnote{The source code of our implementation is shared at
\textcolor{magenta}{http://www.dabi.temple.edu/$\sim$hbling/code/PTAV/ptav.htm}.} a PTAV algorithm needs three parts: a base tracker for $\TRK$, a base verifier for $\VRF$, and the coordination between them. For $\TRK$, we choose the \emph{fast discriminative scale space tracking} (fDSST) algorithm~\cite{danelljan2016discriminative}, which is correlation filter-based and runs efficiently by itself. For $\VRF$, we choose the Siamese networks~\cite{chopra2005learning} for verification similar to~\cite{tao2016siamese}. For coordination, $\TRK$ sends results to $\VRF$ at a certain frequency that allows enough time for verification. On the verifier side, when an unreliable result is found, $\VF$ performs detection and sends the detected result to $\TRK$ for correction. The framework is illustrated in Fig.~\ref{detail_PTV} and detailed in Sec.~3.

The proposed PTAV algorithm is evaluated thoroughly on several benchmarks including OTB2013 \cite{wu2013online}, OTB2015 \cite{wu2015object}, TC128 \cite{liang2015encoding} and UAV20L \cite{mueller2016benchmark}. In these experiments, PTAV achieves the best tracking accuracy among all real-time trackers, and in fact even performs even better than many deep learning based solutions.

In summary, our first main contribution is the novel parallel tracking and verifying framework (\ie PTAV). With the framework, we made a second contribution to implement a tracking solution that combines correlation kernel-based tracking and deep learning-based verification. Then, our solution shows very promising results on thorough experiments in comparison with state-of-the-arts. Moreover, it is worth noting that PTAV is a very flexible framework and our implementation is far from optimal. We believe there are great rooms for future improvement and generalization.

\section{Related Work}

Visual tracking has been extensively studied with a huge amount of literature. In the following we discuss the most related work and refer readers to \cite{YilmazJS06survey,smeulders2014visual} for recent surveys.

\vspace{0.4em}
\noindent {\bf Related tracking algorithms.} Existing model free visual tracking algorithms are often categorized as either discriminative or generative. Discriminative algorithms usually treat tracking as a classification problem that distinguishes the target from ever-changing background. The classifiers in these methods are learned by, \eg, multiple instance learning (MIL) \cite{babenko2011robust}, compressive sensing \cite{zhang2012real}, P-N learning \cite{kalal2012tracking}, structured output SVMs \cite{hare2016struck}, on-line boosting \cite{grabner2008semi} and so on.
By contrast, generative trackers usually formulate tracking as searching for regions most similar to the target. To this end, various object appearance modeling approaches have been proposed such as incremental subspace learning \cite{ross2008incremental} and sparse representation \cite{mei2009robust,bao2012real,fan2017robust}. Inspired by the powerfulness of deep features in visual recognition~\cite{krizhevsky2012imagenet,simonyan2014very}, some trackers~\cite{ma2015hierarchical,nam2016learning, fan2016sanet,wang2013learning} utilize the deep features for robust object appearance modeling. 

Recently, correlation filters have drawn increasing attention in visual tracking. Bolme {\it et al.} \cite{bolme2010visual} propose a correlation filter tracker via learning the minimum output sum of squared error (MOSSE). Benefitting from the high computational efficiency of correlation filters, MOSSE achieves an amazing speed at hundreds of \emph{fps}. Henriques {\it et al.} \cite{henriques2012exploiting} introduce kernel space into correlation filter and propose a circulant structure with kernel (CSK) method for tracking. CSK is then extended in~\cite{henriques2015high} for further improvement and result in the well-known kernelized correlation filters (KCF) tracker. Later, \cite{danelljan2016discriminative} and \cite{li2014scale} propose to improve KCF by imbedding scale estimations into correlation filters. More efforts on improving KCF can be found in~\cite{liu2015real,ma2015long}, etc.

\vspace{0.4em}
\noindent {\bf Verification in tracking.} The idea of verification is not new in tracking. A notable example is the TLD tracker~\cite{kalal2012tracking}, in which tracking results are validated \emph{per frame} to decide how learning and/or detection shall progress. Similar ideas have been used in other trackers such as \cite{hua2014occlusion,hua2015online}.
Unlike in previous studies, the verification in PTAV runs only on sampled frames. This allows PTAV to use strong verification algorithms (Siamese networks~\cite{chopra2005learning} in our implementation) without worrying about running time efficiency.

Interestingly, tracking by itself can be also formulated as a verification problem that finds the best candidate similar to the target~\cite{tao2016siamese,bertinetto2016fully}. Bertinetto {\it et al.} \cite{bertinetto2016fully} propose a fully-convolutional Siamese networks for visual tracking. In \cite{tao2016siamese}, Tao {\it et al.} formulate tracking as object matching in each frame by Siamese networks. Despite obtaining excellent performance, application of such trackers is severely restricted by the heavy computation for extracting deep features in each frame. By contrast, our solution only treats verification as a way to check and correct the \emph{fast tracker}, and does not run verification per frame.

\section{Parallel Tracking and Verifying (PTAV)}

\subsection{Framework}

A typical PTAV consists of two components: a (fast) tracker $\TK$ and a (reliable) verifier $\VF$. The two components work together toward real-time and high accuracy tracking.
\begin{itemize}
\vspace{-1.55mm}\item \noindent {\bf The tracker $\TK$} is responsible of the ``real-time" requirement of PTAV, and needs to locate the target in each frame. Meanwhile, $\TK$ sends verification request to $\VF$ from time to time (though not every frame), and responds to feedback from $\VF$ by adjusting tracking or updating models. To avoid heavy computation, $\TK$ maintains a buffer of tracking information (\eg, intermediate status) in recent frames to facilitate fast tracing back when needed.

\vspace{-1.55mm}\item \noindent {\bf The verifier $\VF$} is employed to pursue the ``high accuracy" requirement of PTAV. Up on receiving a request from $\TK$, $\VF$ tries the best to first validate the tracking result (\eg comparing it with the initialization), and then provide feedback to $\TK$.
\end{itemize}

In PTAV, $\TK$ and $\VF$ run in parallel on two different threads with necessary interactions, as illustrated in Fig.~\ref{detail_PTV}. The tracker $\TK$ and verifier $\VF$ are initialized in the first frame. After that, $\TK$ starts to process each arriving frame and generates the result (pink solid dot in Figure \ref{detail_PTV}). In the meantime, $\VF$ validates the tracking result every several frames. Because tracking is much faster verifying, $\TK$ and $\VF$ work asynchronously. Such mechanism allows PTAV to tolerate temporal tracking drift (\eg, at frame 380 in Figure \ref{detail_PTV}), which will be corrected later by $\VF$. When finding the tracking result unreliable, $\VF$ searches the correct answer from a local region and sends it to $\TK$. Upon the receipt of such feedback, $\TK$ stops current tracking job and traces back to resume tracking with the correction provided by $\VF$.

It is worth noting that PTAV is a very flexible framework, and some important designing choices are following. (1) The choices of base algorithms for $\TK$ and $\VF$ may depend on applications and available computational resources. In addition, in practice one may use more than one verifiers or even base trackers. (2) The response of $\TK$ to the feedback, either positive or negative, from $\VF$ can be largely designed. (3) The correction of unreliable tracking results can be implemented in many ways, and the correction can even be conducted purely by $\TK$ (\ie, including target detection). (4) $\TK$ has various ways to use pre-computed archived information for speeding up. Algorithm \ref{PTAV_alg} summarizes the general PTAV framework.

\subsection{PTAV Implementation}

\subsubsection{Tracking}

We choose the fDSST tracker \cite{danelljan2016discriminative} as the base of the tracker $\TK$ in PTAV. As a discriminative correlation filter-based tracker, fDSST learns a model on an image patch $f$ with a $d$-dimension feature vector. Let $f^{l}$ denote the feature along the $l$-th dimension, $l\in{1,2,\cdots,d}$. The objective is to learn the optimal correlation filter $h$, consisting of one filter $h^{l}$ per feature dimension, by minimizing the cost function
\begin{equation}
\epsilon(f) = \Big\|{\sum\nolimits_{l=1}^{d}{h^{l}*f^{l}}-g} \Big\|^{2} + \lambda\sum\nolimits_{l=1}^{d}{\parallel{h^{l}}\parallel^{2}}\label{eq1}
\end{equation}
where $g$ represents the desired correlation output associated with the training patch $f$, $\lambda$ ($\lambda\geqslant{0}$) denotes a tradeoff parameter, and $*$ is circular convolution operation.

Using the fast Fourier transformation (FFT), the solution of Eq. (\ref{eq1}) can be efficiently obtained with
\begin{equation}
H^{l} = \frac{\bar{G}F^{l}}{\sum\nolimits_{k=1}^{d}{\bar{F^{k}}{F^{k}}+\lambda}}, \;\;\;\; l = 1,2,\cdots,d \label{eq2}
\end{equation}
where the capital letters in Eq. (\ref{eq2}) represent the discrete Fourier transform (DFT) of the corresponding quantities, and the bar $(\bar{\cdot})$ indicates complex conjugation.

\begin{algorithm}[!t]
\small
\caption{Parallel Tracking and Verifying (PTAV)}\label{PTAV_alg}
    Initialize the tracking thread for tracker $\TK$\;
    Initialize the verifying thread for verifier $\VF$\;
    Run $\TK$ (Alg.~\ref{PTAV_T}) and $\VF$ (Alg.~\ref{PTAV_V}) till the end of tracking\;
\end{algorithm}

\begin{algorithm}[!t]\small
\caption{Tracking Thread $\TK$}\label{PTAV_T}
    \While {Current frame is valid}
    {
        \eIf {received a message $s$ from $\VF$}
        {
            \eIf{verification passed}
            {
                Update tracking model (optional)\;
            }
            {
                Correct tracking\;
                Trace back and reset current frame\;
                Resume tracking\;
            }
        }
        {
            Tracking on the current frame\;
            \If{time for verification}
            {
                Send the current result to $\VF$ to verify;
            }
        }
        Current frame $\leftarrow$ next frame\;
    }
\end{algorithm}

\begin{algorithm}[!t]\small
\caption{Verifying Thread $\VF$}\label{PTAV_V}
    \While{not ended}
    {
        \If {received request from $\TK$}
        {
            Verifying the tracking result\;
            \If{verification failed}
            {
                Provide correction information\;
            }
            Send verification result $s$ to $\TK$\;
        }
    }
\end{algorithm}

An optimal filter can be derived by minimizing the output error over all training samples \cite{kiani2013multi}. Nevertheless, this requires solving a $d\times{d}$ linear system of equations per pixel, leading to high computation. For efficiency, a simple linear update strategy is applied to the numerator $A_{t}^{l}$ and denominator $B_{t}$ of $H_{t}^{l}$ by
\begin{equation}
\begin{array}{lcl}
    A_{t}^{l} & = & (1-\eta)A_{t-1}^{l} + \eta{\bar{G}_{t}F_{t}^{l}} \\
    B_{t}     & = & (1-\eta)B_{t-1} + \eta\sum\nolimits_{k=1}^{d}{\bar{F_{t}^{k}}F_{t}^{k}}  \label{eq3}
\end{array}
\end{equation}
where $\eta$ is the learning rate. The responding scores $y$ for a new image patch $z$ can be computed by
\begin{equation}
y=\mathcal{F}^{-1}	\bigg\{ \frac{\sum_{l=1}^{d}{\bar{A^{l}}Z^{l}}}{B+\lambda} \bigg\}
\end{equation}
The position of target object is determined by the location of maximal value of $y$.

To adapt the tracker to scale variation, a scale filter is adopted to estimate the scale of target. In addition, to further decrease computation, principal component analysis (PCA) is utilized to reduce the dimension of feature vector. For more details, readers are referred to \cite{danelljan2016discriminative}.

Unlike in~\cite{danelljan2016discriminative}, in our implementation the tracker $\TK$ stores all intermediate results (\eg $H_{t}^{l}$ in each frame $t$) since sending out last verification request to ensure fast tracing back.  To validate the tracking result, $\TK$ sends the verification results every $V$ frames, where $V$ is the dynamically adjustable verification interval as described later.

\subsubsection{Verifying}
We adopt the siamese networks~\cite{chopra2005learning} to develop the verifier $\VF$. The siamese networks\footnote{Due to page limitation, we refer readers to the supplementary material for detailed architecture of the siamese networks and its training process.} contain two branches of CNNs, and process two inputs separately. In this work, we borrow the architecture from VGGNet~\cite{simonyan2014very} for the CNNs, but with an additional region pooling layer \cite{girshick2015fast}. This is because, for detection, $\VF$ needs to process multiple regions in an image, from which the candidate most similar to the target is selected to be result. As a result, region pooling enables us to simultaneously process a set of regions in an image.

\begin{figure}[!t]
\centering
\begin{tabular}{@{}C{8.2cm}@{}}
\includegraphics[width=8.15cm]{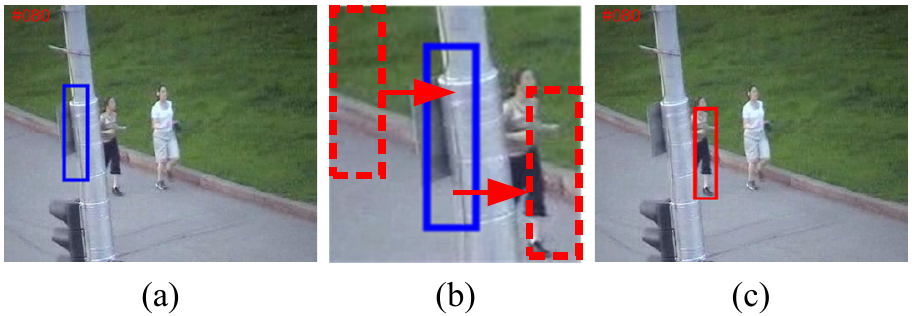}\\
\end{tabular}
\caption{Detection based on verification. When finding an unreliable tracking result (showing in \textcolor{blue}{blue} in (a)), the verifier $\VF$ detects the target in a local region ( shown in (b)). The dashed \textcolor{red}{red} rectangles in (b) represent object candidates generated by sliding window. The \textcolor{red}{red} rectangle in (c) is the detection result.}
\label{detection}
\end{figure}

Given the tracking result from $\TRK$, if its verification score is lower than a threshold $\tau_{1}$, $\VF$ will treat it as a tracking failure. In this case, $\VF$ needs detect target, again using the siamese networks. Unlike for verification, detection requires to verify multiple image patches in a local region\footnote{The local region is a square of size $\beta(\mathrm{w}^{2}+\mathrm{h}^{2})^{\frac{1}{2}}$ centered at the location of the tracking result in this validation frame, where $\mathrm{w}$ and $\mathrm{h}$ are the width and height of the tracking result, and $\beta$ controls the scale.} and finds the best one. Thanks to the region pooling layer, these candidates can be simultaneously processed in only one pass, resulting in significant reduction in computation. Let $\{c_{i}\}_{i=1}^{N}$ denote the candidate set generated by sliding window, and the detection result $\widehat{c}$ is determined by
\begin{equation}
\widehat{c} = \argmax_{c_{i}}{\nu(x_{obj}, c_{i})}, \;\;\;\; i =1,2,\cdots,N
\end{equation}
where $\nu(x_{obj}, c_{i})$ returns the verification score between the tracking target $x_{obj}$ and the candidate $c_i$.

After obtaining detection result, we determine whether or not take it to be an alternative for tracking result based on its verification score. If detection result is unreliable (e.g., the verification score for detection result is less than a threshold $\tau_{2}$), we do not change the tracking result. Instead, we decrease the verifying interval $V$, and increase the size of local region to search for the target. The process repeats until we find a reliable detection result. Then we restore verification interval and the size of the searching region. Figure \ref{detection} shows the detection process.

\begin{figure*}[!t]
\centering
\begin{tabular}{@{}C{8.8cm}@{}@{}C{8.8cm}@{}@{}}
\includegraphics[width=4.5cm]{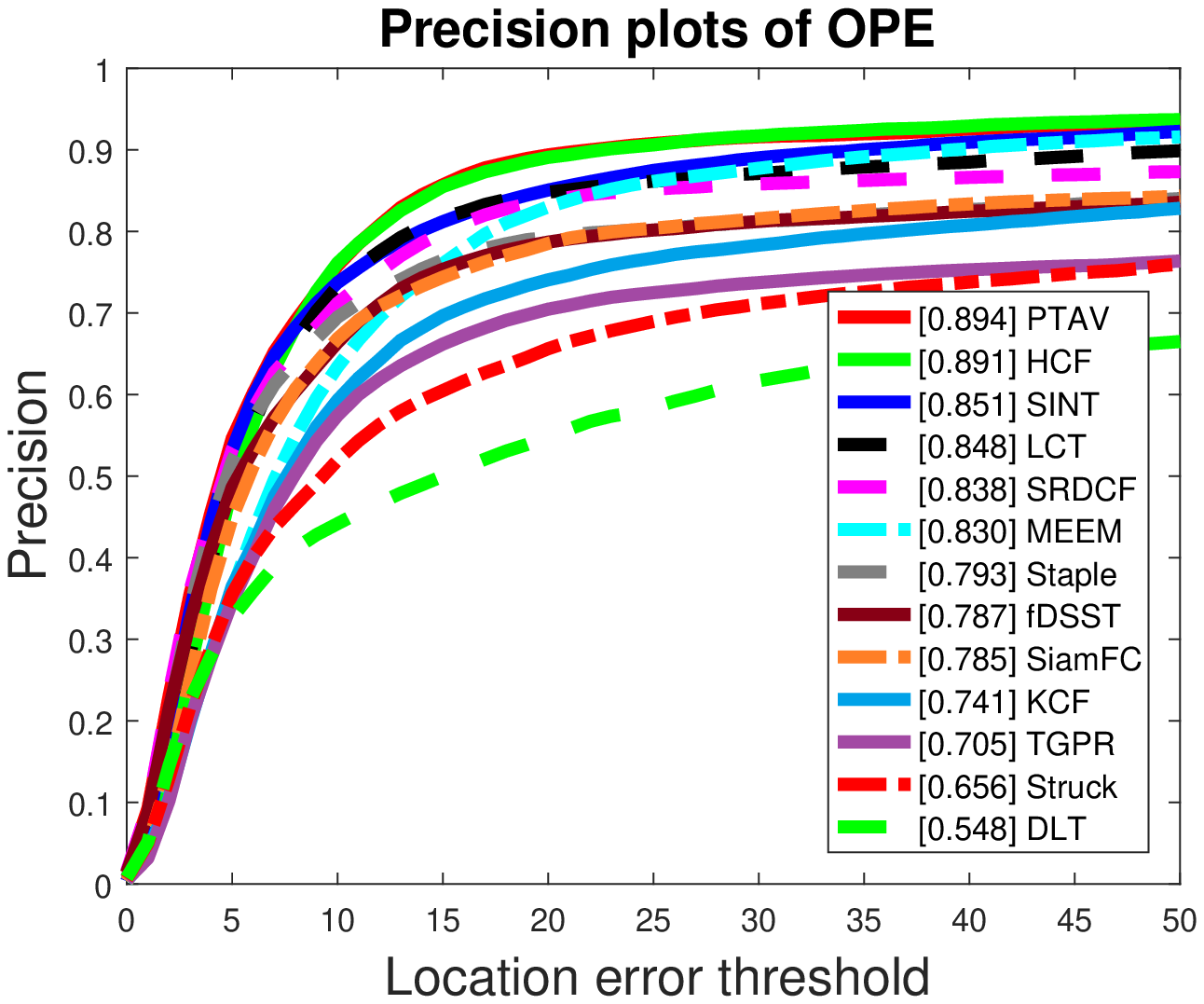}\includegraphics[width=4.5cm]{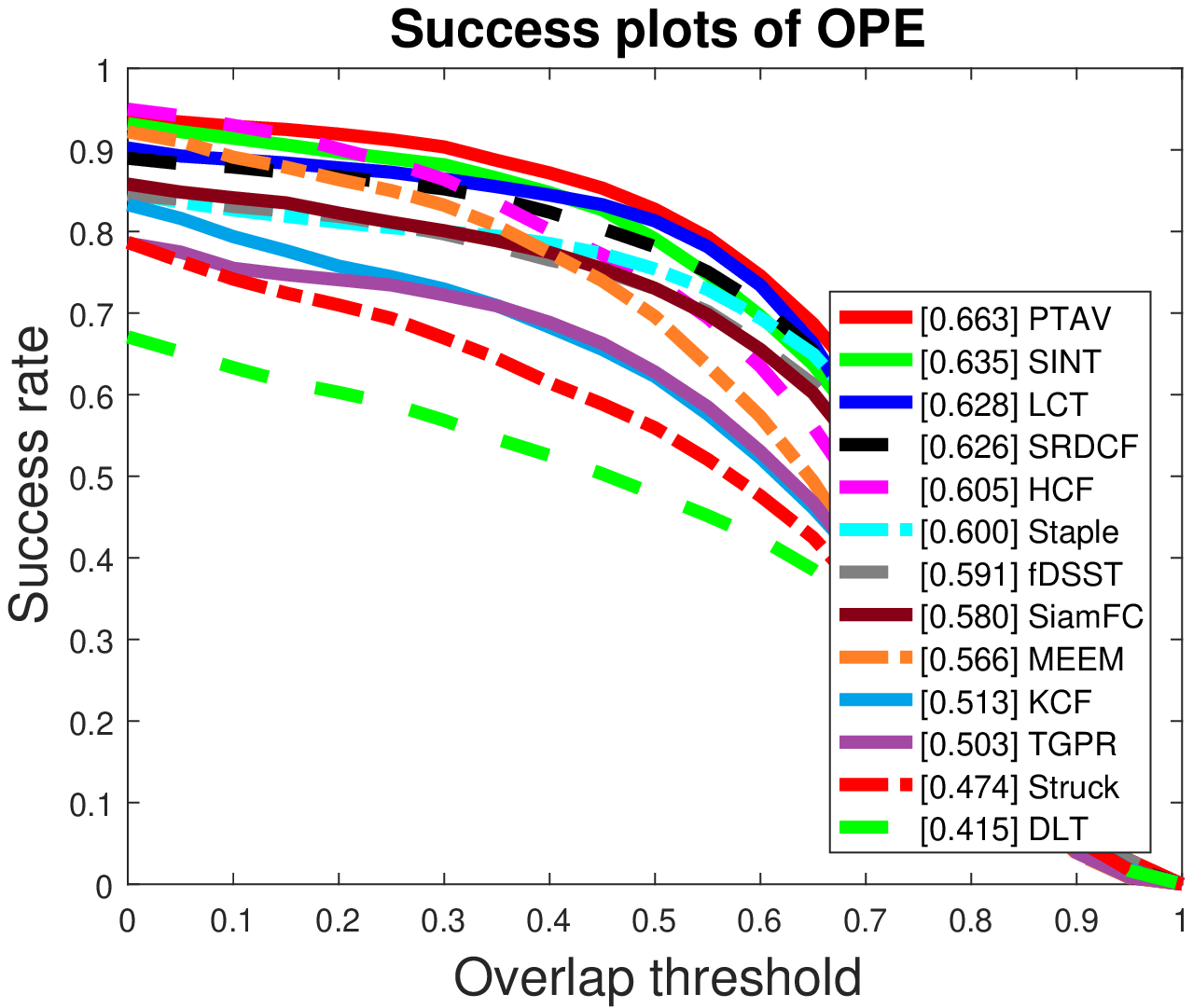} & \includegraphics[width=4.5cm]{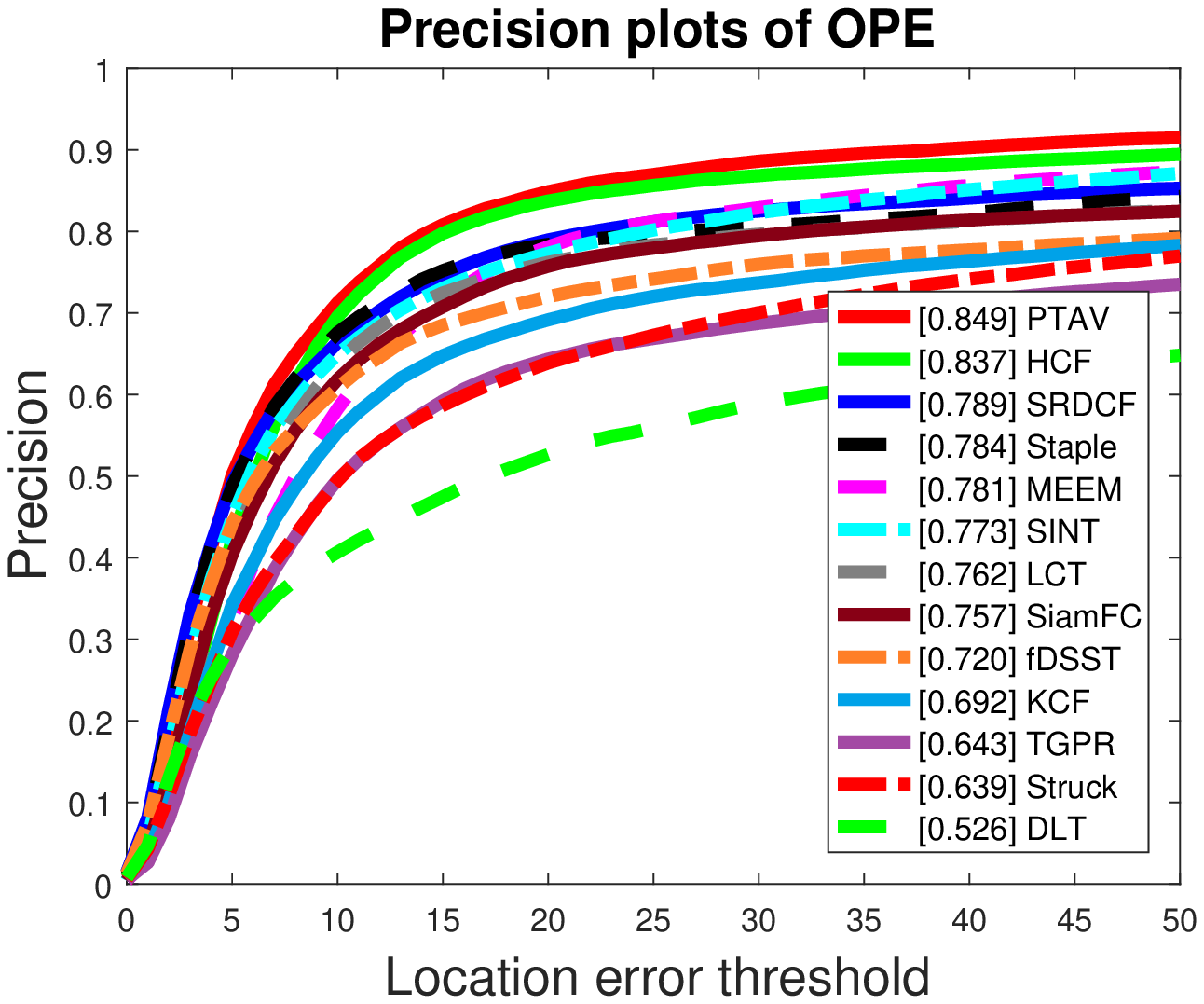}\includegraphics[width=4.5cm]{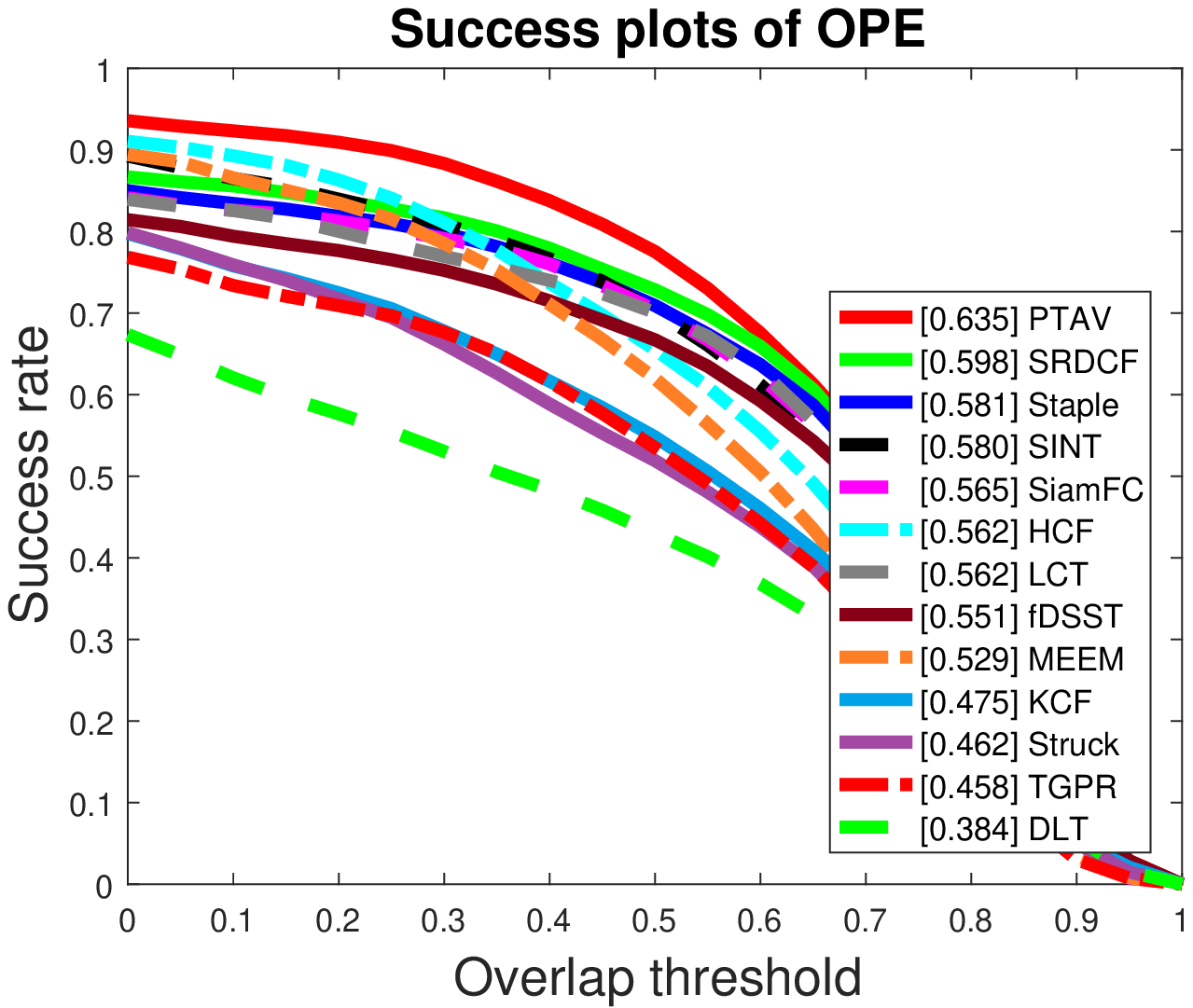}\\
 {\small (a) Comparisons on OTB2013} & {\small (b) Comparisons on OTB2015} \\
\end{tabular}
\caption{Comparison on OTB2013 and OTB2015 using distance precision rate (DPR) and overlap success rate (OSR).}
\label{comparison_OTB}
\end{figure*}
\renewcommand\arraystretch{1.2}
\begin{table*}[!t]\footnotesize
  \centering
  \caption{Comparisons with state-of-the-art tracking methods on OTB2013 \cite{wu2013online} and OTB2015 \cite{wu2015object}. Our PTAV outperforms existing approaches in distance precision rate (DPR) at a threshold of 20 pixels and overlap success rate (OSR) at an overlap threshold of 0.5.}
    \begin{tabular}{|@{}C{1.3cm}@{}|@{}C{2.1cm}@{}|@{}C{1.1cm}@{}|@{}C{1.0cm}@{}C{1.1cm}@{}C{1.0cm}@{}C{1.1cm}@{}|@{}C{1.1cm}@{}C{1.0cm}@{}C{1.0cm}@{}C{1.1cm}@{}C{1.0cm}@{}|@{}C{1.1cm}@{}C{1.0cm}@{}C{1.1cm}@{}|}
    \hline
    \multirow{2}[0]{*}{} & \multirow{2}[0]{*}{} & \multirow{2}[0]{*}{} & \multicolumn{4}{c|}{Deep trackers} & \multicolumn{5}{c|}{Correlation-filters based trackers} & \multicolumn{3}{c|}{Representative trackers} \\
    \cline{4-15}
            &         &     PTAV (Ours)    & HCF \cite{ma2015hierarchical}     & SINT \cite{tao2016siamese}    & DLT \cite{wang2013learning} & SiamFC \cite{bertinetto2016fully}     & SRDCF \cite{danelljan2015learning}   & Staple \cite{bertinetto2016staple}  & LCT \cite{ma2015long}    & fDSST \cite{danelljan2016discriminative}   & KCF \cite{henriques2015high}     & MEEM \cite{zhang2014meem}    & TGPR \cite{gao2014transfer}    & Struck \cite{hare2016struck} \\
            \cline{1-15}
    \multirow{3}[0]{*}{OTB2013} & DPR (\%) & 89.4 & 89.1 & 85.1    & 54.8 & 78.5    & 83.8    & 79.3    & 84.8    & 78.7    & 74.1    & 83      & 70.5    & 65.6 \\
            & OSR (\%) & 82.7 & 74      & 79.1    & 47.8 & 74.0    & 78.2    & 75.4    & 81.2 & 74.7    & 62.2    & 69.6    & 62.8    & 55.9 \\
            \cline{2-15}
            & Speed (\it{fps}) & 27      & 11      & 3       & 9  & 46     & 4       & 45      & 27      & 54 & 245 & 21      & 1       & 10 \\
            \cline{1-15}
    \multirow{3}[0]{*}{OTB2015} & DPR (\%) & 84.9 & 83.7 & 77.3    & 52.6  & 75.7  & 78.9    & 78.4    & 76.2    & 72      & 69.2    & 78.1    & 64.3    & 63.9 \\
            & OSR (\%) & 77.6 & 65.6    & 70.3    & 43  & 70.9    & 72.9 & 70.9    & 70.1    & 67.6    & 54.8    & 62.2    & 53.5    & 51.6 \\
            \cline{2-15}
            & Speed (\it{fps}) & 25      & 10      & 2       & 8   & 43    & 4       & 43      & 25      & 51 & 243 & 21      & 1       & 10 \\
    \hline
    \end{tabular}%
  \label{OTB_table}%
\end{table*}%

\section{Experiments}
\subsection{Implementation details}
Our PTAV is implemented in C++ and its verifier uses Caffe \cite{jia2014caffe} on a single NVIDIA GTX TITAN Z GPU with 6GB memory. The regularization term $\lambda$ in Eq. (\ref{eq1}) is set to 0.01, and the learning rate in Eq. (\ref{eq3}) to 0.025. Other parameters for tracking remain the same as in \cite{danelljan2016discriminative}. The siamese networks for verification are initialized with VGGNet \cite{simonyan2014very} and trained based on the approach in \cite{tao2016siamese}. The verification interval $V$ is initially set to 10. The validation and detection thresholds $\tau_{1}$ and $\tau_{2}$ are set to 1.0 and 1.6, respectively. The parameter $\beta$ is initialized to 1.5, and is adaptively adjusted based on the detection result. If the detection result with $\beta=1.5$ is not reliable, the verifier will increase $\beta$ for a larger searching region. When the new detection becomes faithful, $\beta$ is restored to 1.5.

\subsection{Experiments on OTB2013 and OTB2015}

\noindent {\bf Overall performance.} OTB2013 \cite{wu2013online} and OTB2015 \cite{wu2015object} are two popular tracking benchmarks, which contain 50 and 100 videos, respectively. We evaluate the PTAV on these benchmarks in comparison with 11 state-of-the-art trackers from three typical categories: (\lowercase\expandafter{\romannumeral1}) deep features-based trackers, including SINT \cite{tao2016siamese}, HCF \cite{ma2015hierarchical}, SiamFC \cite{bertinetto2016fully} and DLT \cite{wang2013learning}; (\lowercase\expandafter{\romannumeral2}) correlation filters-based tracking approaches, including fDSST \cite{danelljan2016discriminative}, LCT \cite{ma2015long}, SRDCF \cite{danelljan2015learning}, KCF \cite{henriques2015high} and Staple \cite{bertinetto2016staple}; and (\lowercase\expandafter{\romannumeral3}) other representative tracking methods, including TGPR \cite{gao2014transfer}, MEEM \cite{zhang2014meem} and Struck \cite{hare2016struck}. For SINT \cite{tao2016siamese}, we use its tracking results without optical flow because no optical flow part is provide from the released source code. In PTAV, the fDSST tracker \cite{danelljan2016discriminative} is chosen to be our tracking part, and thus it can be regarded as our baseline. It is worth noting that other tracking algorithms may also be used for tracking part in our PTAV.

Following the protocol in \cite{wu2013online,wu2015object}, we report the results in \emph{one-pass evaluation} (OPE) using \emph{distance precision rate} (DPR) and \emph{overlap success rate} (OSR) as shown in Fig.~\ref{comparison_OTB}. Overall, PTAV performs favorably against all other state-of-the-art trackers on both datasets. In addition, we present quantitative comparison of DPR at 20 pixels, OVR at 0.5, and speed in Table~\ref{OTB_table}. It shows that PTAV outperforms other state-of-the-art trackers in both rates. On OTB2015, our tracker achieves a DPR of 84.9\% and an OVR of 77.6\%. Though the HCF \cite{ma2015hierarchical} utilizes deep features to represent object appearance, our approach performs better compared with its DPR of 83.7\% and OSR of 65.6\%. Besides, owing to the adoption of parallel framework, PTAV (27 fps) is more than twice faster than HCF (10 fps). Compared with SINT \cite{tao2016siamese}, PTAV improves DPR from 77.3\% to 84.9\% and OSR from 70.3\% to 77.6\%. In addition, PTAV runs at real-time while SINT does not. Compared with the baseline fDSST \cite{danelljan2016discriminative}, PTAV achieves significant improvements on both DPR (12.9\%) and OSR (10.0\%).

\renewcommand\arraystretch{1.0}
\begin{table*}[!t]\small
  \centering
  \caption{Average precision and success scores of PTAV and other five top trackers on different attributes: background cluttered
(BC), deformation (DEF), fast motion (FM), in-plane rotation (IPR), illumination variation (IV), low resolution (LR), motion blur (MB), occlusion (OCC), out-of-plane rotation (OPR), out-of-view (OV) and scale variation (SV).}
    \begin{tabular}{@{}C{1.2cm}@{}|
    @{}C{.9cm}@{}@{}C{1.23cm}@{}@{}C{1.52cm}@{}@{}C{1.52cm}@{}@{}C{1.54cm}@{}@{}C{1.35cm}@{} | @{}C{.9cm}@{}@{}C{1.23cm}@{}@{}C{1.52cm}@{}@{}C{1.52cm}@{}@{}C{1.54cm}@{}@{}C{1.35cm}@{} }    
    \hline
          & \multicolumn{6}{c|}{Distance precision rate (\%) on eleven attributes} & \multicolumn{6}{c}{Overlap success rate (\%) on eleven attributes} \\
    \hline
    Attribute & PTAV  & HCF \cite{ma2015hierarchical}   & SRDCF \cite{danelljan2015learning} & Staple \cite{bertinetto2016staple} & MEEM\cite{zhang2014meem}  & SINT \cite{tao2016siamese}  &
    PTAV  & HCF \cite{ma2015hierarchical}   & SRDCF \cite{danelljan2015learning} & Staple \cite{bertinetto2016staple} & MEEM \cite{zhang2014meem}  & SINT \cite{tao2016siamese} \\
    \hline
    BC    &  87.9  & 84.7  & 77.6  & 77.0  & 75.1  & 75.1  & 64.9  & 58.7  & 58.0  & 57.4  & 52.1  & 56.7 \\
    DEF   &  81.3  & 79.1  & 73.4  & 74.8  & 75.4  & 75.0  & 59.7  & 53.0  & 54.4  & 55.4  & 48.9  & 55.5 \\
    FM    &  77.7  & 79.7  & 76.8  & 70.3  & 73.4  & 72.5  & 60.8  & 55.5  & 59.9  & 54.1  & 52.8  & 55.7 \\
    IPR   &  83.0  & 85.4  & 74.5  & 77.0  & 79.3  & 81.1  & 60.7  & 55.9  & 54.4  & 55.2  & 52.8  & 58.5 \\
    IV    &  86.0  & 81.7  & 79.2  & 79.1  & 74.0  & 80.9  & 64.3  & 54.0  & 61.3  & 59.8  & 51.7  & 61.8 \\
    LR    &  78.9  & 78.7  & 63.1  & 60.9  & 60.5  & 78.8  & 56.3  & 42.4  & 48.0  & 41.1  & 35.5  & 53.9 \\
    MB    &  81.0  & 79.7  & 78.2  & 72.6  & 72.1  & 72.8  & 62.9  & 57.3  & 61.0  & 55.8  & 54.3  & 57.4 \\
    OCC   &  83.2  & 76.7  & 73.5  & 72.6  & 74.1  & 73.1  & 62.3  & 52.5  & 55.9  & 54.8  & 50.3  & 55.8 \\
    OPR   &  82.8  & 81.0  & 74.6  & 74.2  & 79.8  & 79.4  & 61.1  & 53.7  & 55.3  & 53.8  & 52.8  & 58.6 \\
    OV    &  73.6  & 67.7  & 59.7  & 66.1  & 68.3  & 72.5  & 57.0  & 47.4  & 46.0  & 48.1  & 48.4  & 55.9 \\
    SV    &  79.7  & 80.2  & 74.9  & 73.1  & 74.0  & 74.2  & 59.0  & 48.8  & 56.5  & 52.9  & 47.3  & 55.8 \\
    \hline
    Overall &  84.9  & 83.7  & 78.9  & 78.4  & 78.1  & 77.3  & 63.5  & 56.2  & 59.8  & 58.1  & 52.9  & 58.0 \\
    \hline
    \end{tabular}%
  \label{OTB2015_attribute}%
\end{table*}%

\begin{figure*}[!t]
\centering
\begin{tabular}{@{\hspace{.0mm}}c@{\hspace{1.95mm}} @{\hspace{.0mm}}c@{\hspace{.0mm}}}
\includegraphics[width=2.05cm]{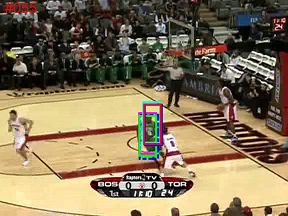} \includegraphics[width=2.05cm]{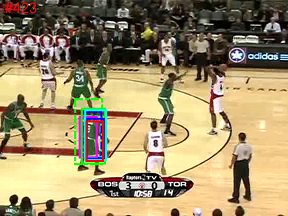} \includegraphics[width=2.05cm]{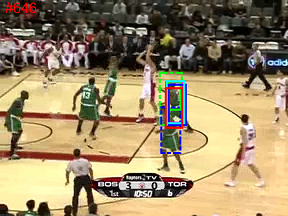} \includegraphics[width=2.05cm]{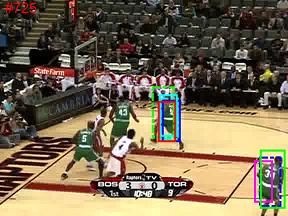} & \includegraphics[width=2.05cm]{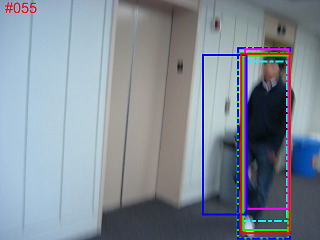} \includegraphics[width=2.05cm]{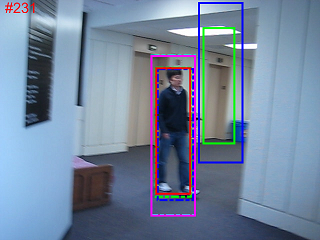} \includegraphics[width=2.05cm]{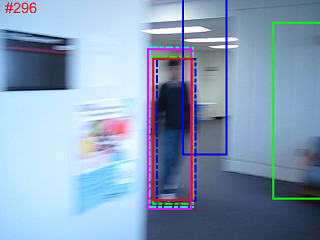} \includegraphics[width=2.05cm]{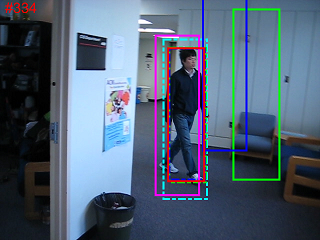} \\
\includegraphics[width=2.05cm]{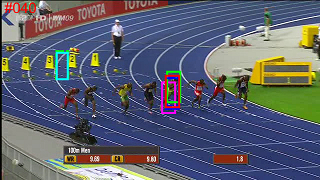} \includegraphics[width=2.05cm]{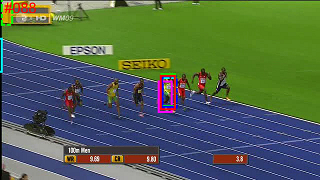} \includegraphics[width=2.05cm]{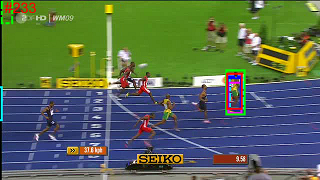} \includegraphics[width=2.05cm]{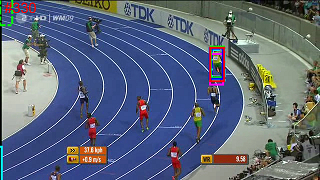} & \includegraphics[width=2.05cm]{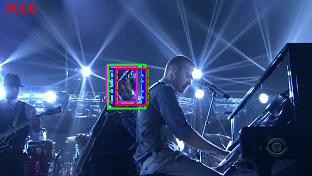} \includegraphics[width=2.05cm]{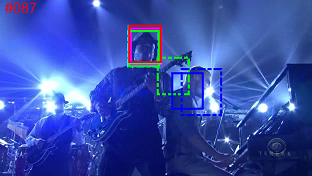} \includegraphics[width=2.05cm]{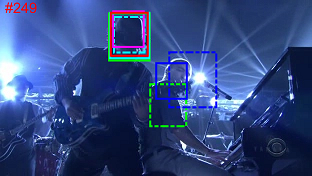} \includegraphics[width=2.05cm]{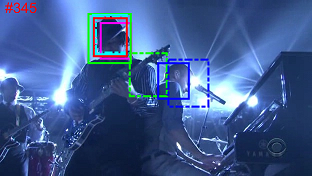} \\
\includegraphics[width=2.05cm]{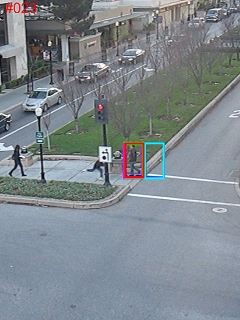} \includegraphics[width=2.05cm]{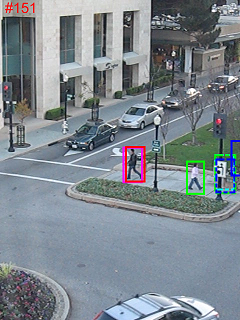} \includegraphics[width=2.05cm]{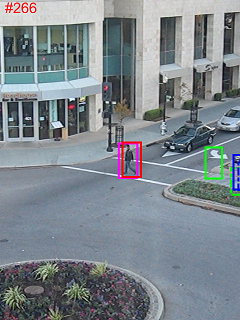} \includegraphics[width=2.05cm]{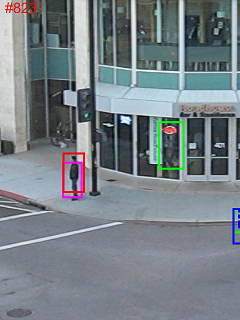} & \includegraphics[width=2.05cm]{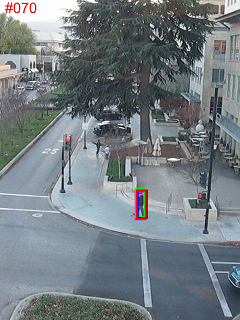} \includegraphics[width=2.05cm]{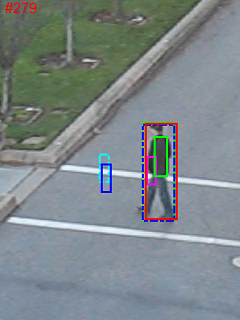} \includegraphics[width=2.05cm]{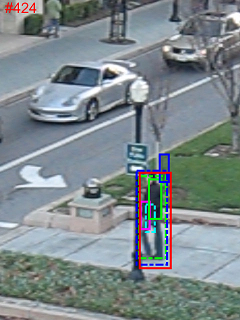} \includegraphics[width=2.05cm]{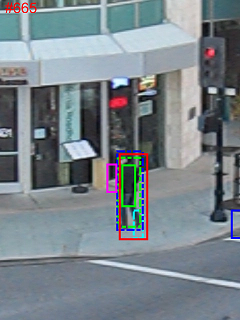} \\
\includegraphics[width=2.05cm]{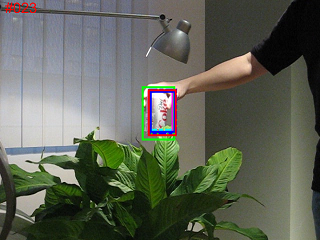} \includegraphics[width=2.05cm]{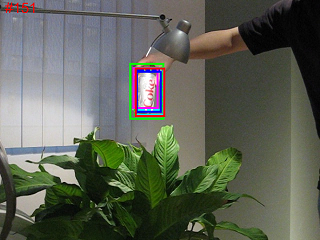} \includegraphics[width=2.05cm]{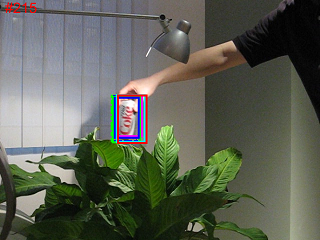} \includegraphics[width=2.05cm]{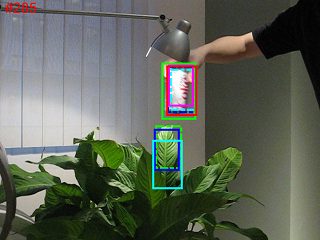} & \includegraphics[width=2.05cm]{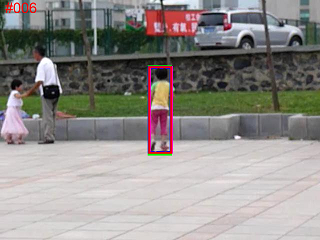} \includegraphics[width=2.05cm]{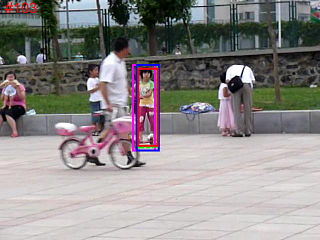} \includegraphics[width=2.05cm]{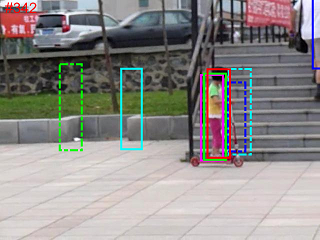} \includegraphics[width=2.05cm]{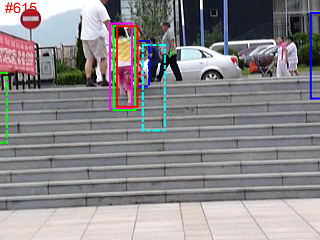} \\
\includegraphics[width=2.05cm]{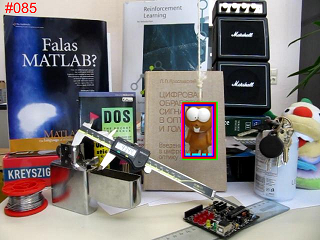} \includegraphics[width=2.05cm]{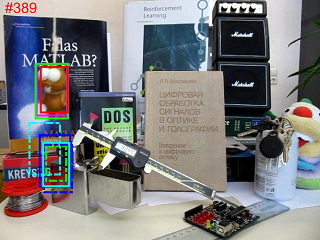} \includegraphics[width=2.05cm]{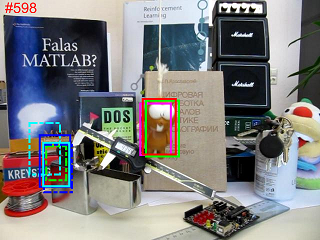} \includegraphics[width=2.05cm]{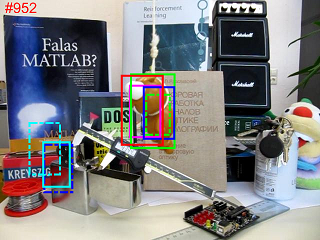} & \includegraphics[width=2.05cm]{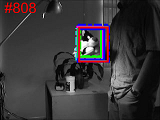} \includegraphics[width=2.05cm]{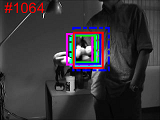} \includegraphics[width=2.05cm]{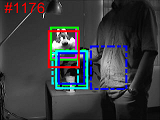} \includegraphics[width=2.05cm]{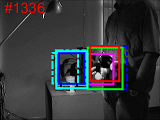} \\
\includegraphics[width=2.05cm]{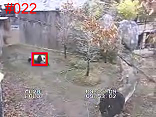} \includegraphics[width=2.05cm]{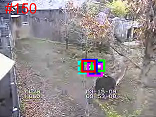} \includegraphics[width=2.05cm]{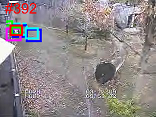} \includegraphics[width=2.05cm]{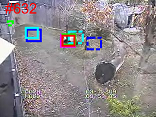} & \includegraphics[width=2.05cm,height=1.49358cm]{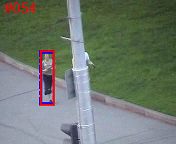} \includegraphics[width=2.05cm,height=1.49358cm]{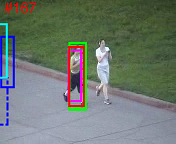} \includegraphics[width=2.05cm,height=1.49358cm]{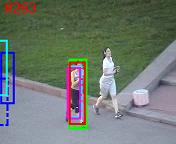} \includegraphics[width=2.05cm,height=1.49358cm]{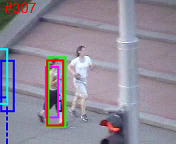} \\
\multicolumn{2}{c}{\includegraphics[width=14.cm]{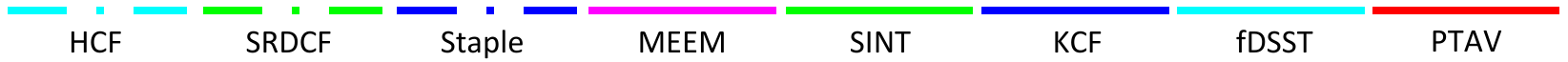}}\\
\end{tabular}
\caption{Qualitative evaluation of the proposed algorithm and other seven state-of-the-art trackers on twelve sequences (from left to right and top to bottom: {\it Basketball}, {\it BlurBody}, {\it Bolt}, {\it Shaking}, {\it Human3}, {\it Human6}, {\it Coke}, {\it Girl2}, {\it Lemming}, {\it Sylvester}, {\it Panda}, and {\it Jogging-1}.)}
\label{qua_res}
\end{figure*}

\noindent {\bf Attribute-based evaluation.} We further analyze the performance of PTAV under different attributes in OTB2015~\cite{wu2015object}. Table~\ref{OTB2015_attribute} shows the comparison of PTAV with other top five tracking algorithms on these eleven attributes. In terms of distance precision rates (DPR), PTAV achieves the best results under 8 out of 11 attributes. For the rest three (FM, IPR and SV), PTAV obtains competitive performances. Compared with other deep learning-based trackers~\cite{ma2015hierarchical,tao2016siamese}, PTAV can better locate the target object in videos. On the other hand, PTAV achieves the best results of overlap success rates (OVR) under all 11 attributes. Compared with correlation filters-based trackers~\cite{danelljan2015learning,bertinetto2016staple} and MEEM~\cite{zhang2014meem}, PTAV performs more robust under occlusion, background cluttered and low resolution with the help of cooperation between tracker and verifier.

\noindent {\bf Qualitative evaluation.} Figure \ref{qua_res} summarizes qualitative comparisons of PTAV with seven state-of-the-art trackers (HCF \cite{ma2015hierarchical}, SRDCF \cite{danelljan2015learning}, Staple \cite{bertinetto2016staple}, MEEM \cite{zhang2014meem}, SINT \cite{tao2016siamese}, KCF \cite{henriques2012exploiting} and fDSST \cite{danelljan2016discriminative}) on twelve sequences sampled form OTB2015 \cite{wu2015object}. The correlation filters-based trackers (KCF \cite{henriques2015high}, SRDCF \cite{danelljan2015learning}, fDSST \cite{danelljan2016discriminative} and Staple \cite{bertinetto2016staple}) perform well in sequences with deformation, illumination variation and partial occlusion ({\it Basketball}, {\it Bolt}, {\it Shaking} and {\it Panda}). However, when full occlusion happens ({\it Coke} and {\it Jogging-1}), they are prone to lose the target. HCF \cite{ma2015hierarchical} uses deep features to represent object appearance, and can deal with these cases to some degree. Nevertheless, it still fails when occlusion happens with other situations such as deformation and rotation ({\it Girl2}, {\it Human3}, {\it Human6}, {\it Sylvester} and {\it Lemming}).

Compared with these trackers, PTAV locate the target object more reliably. Even when experiencing a short drift, the verifier in PTAV can sense the drift and then detect the correct target for subsequent tracking. SINT \cite{tao2016siamese} deals well with occlusion thanks to its capability in re-locating the target. However, it meets problems when motion blur occurs ({\it BlurBody}), which causes serious change in it extracted features. Different from SINT, PTAV uses a correlation filters based method for its tracking part, which works well for motion blur. MEEM \cite{zhang2014meem} uses multiple classifier to track the target and works well in most cases. However, it may lose the target in presences of heavy occlusion and scale variations (\eg, {\it Human6}).

\begin{figure}[!t]
\centering
\includegraphics[width=4.34cm]{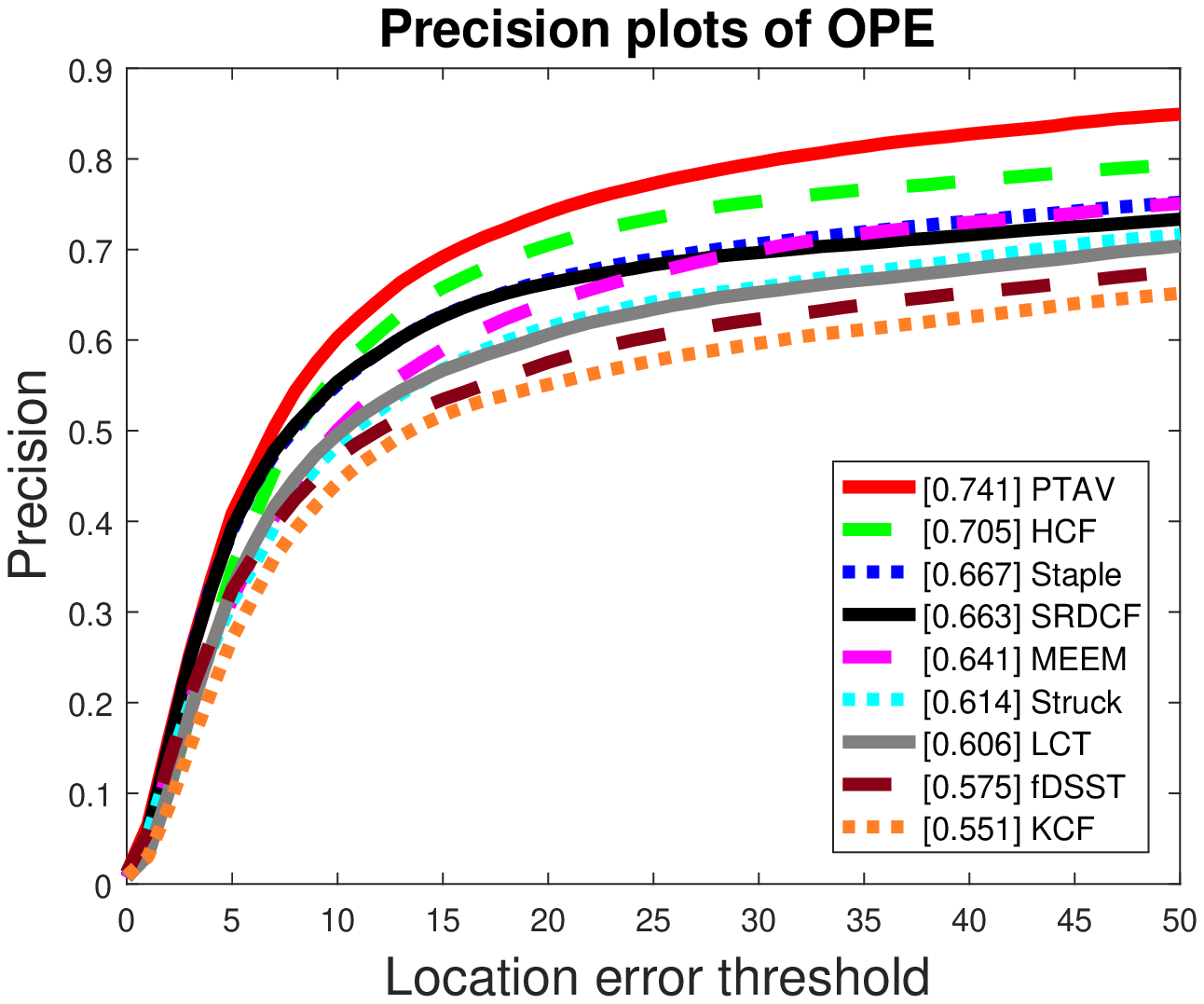}\includegraphics[width=4.34cm]{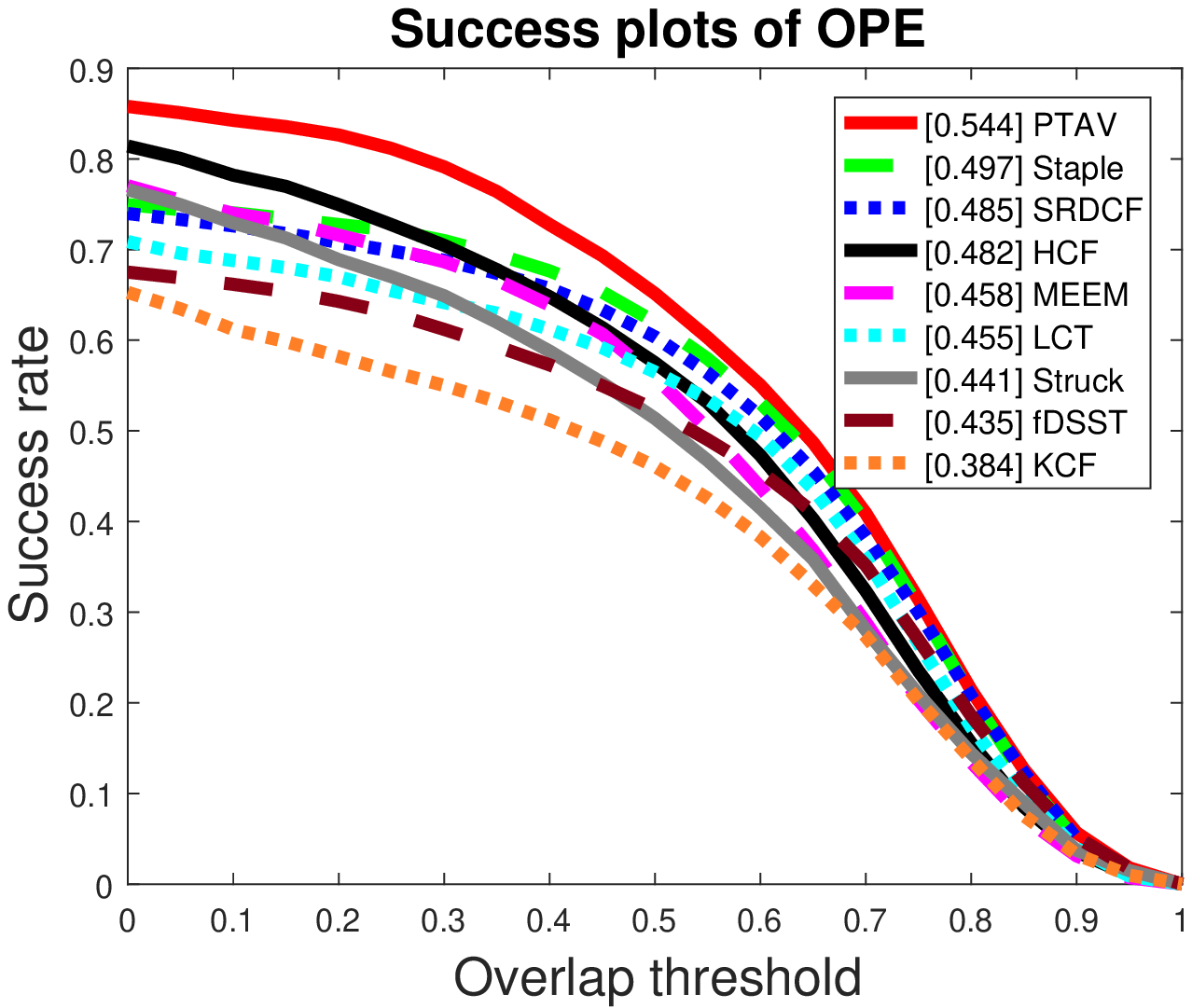}
\caption{Comparison on TC128 using DPR and OSR.}
\label{comparison_TC}
\end{figure}

\begin{figure}[!t]
\centering
\includegraphics[width=4.34cm]{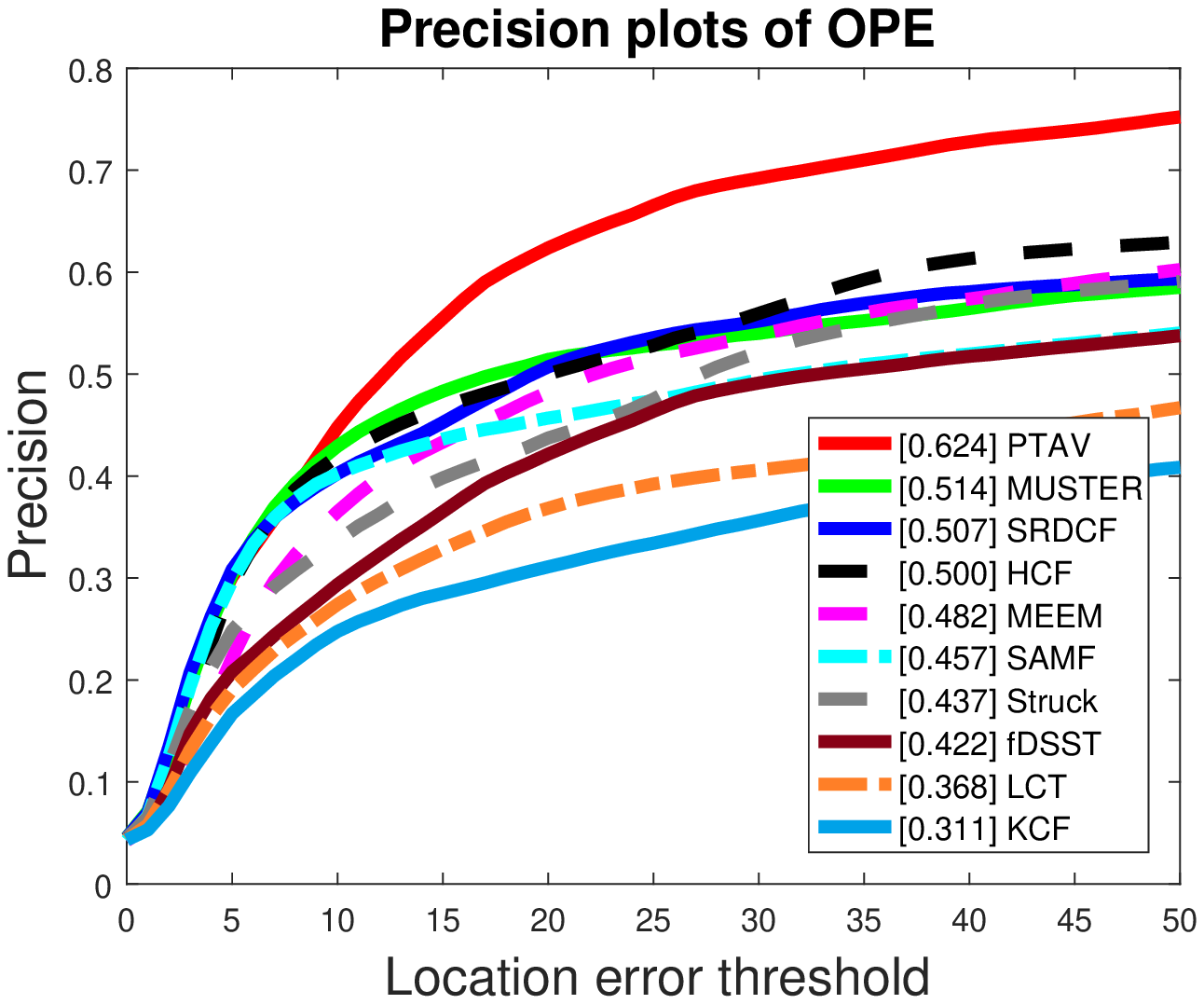}\includegraphics[width=4.34cm]{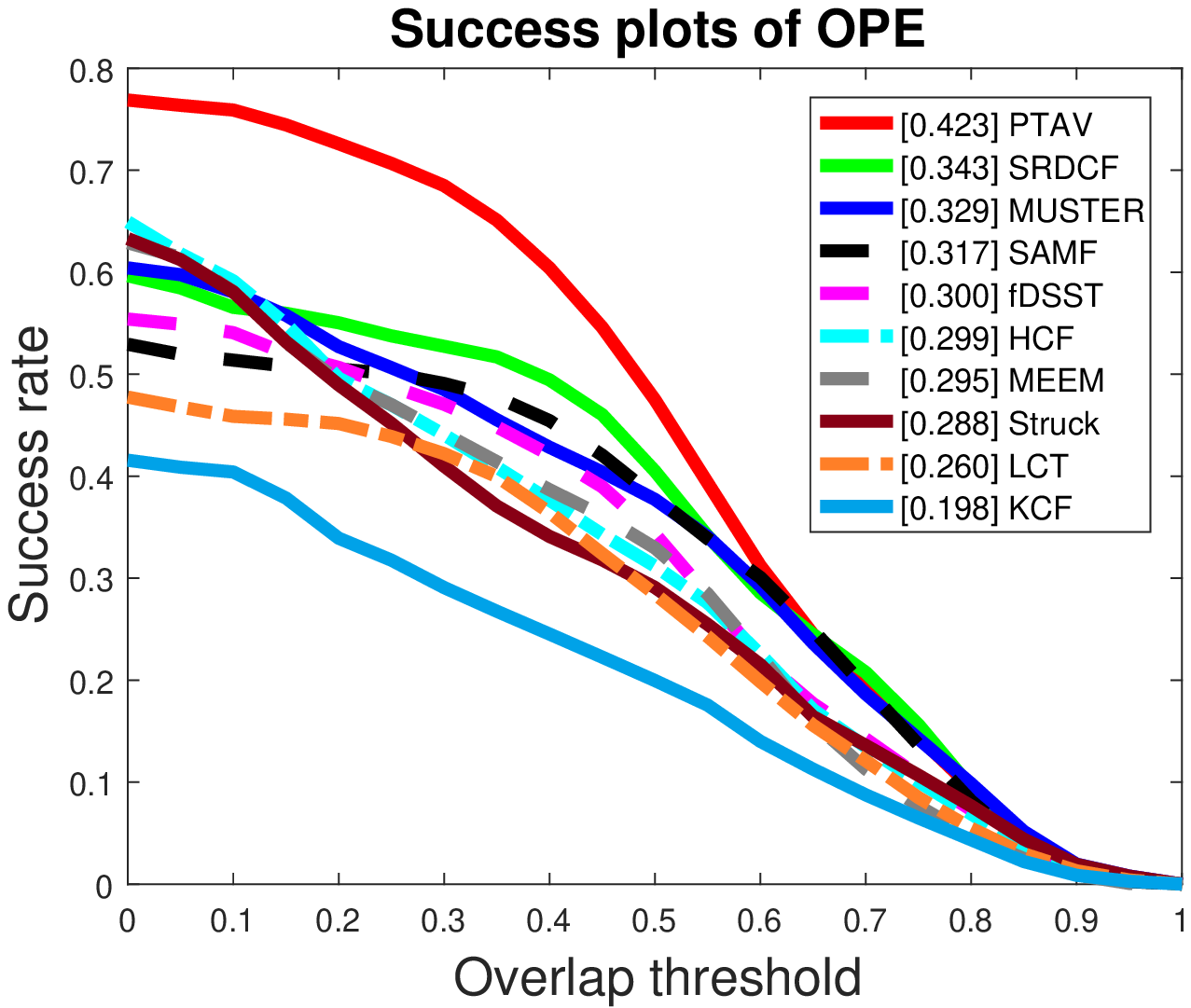}
\caption{Comparison on UAV20L using DPR and OSR.}
\label{comparison_UAV20L}
\end{figure}

\subsection{Experiment on TC128}
For experiments on the TC128 \cite{liang2015encoding} dataset containing 128 videos, our PTAV runs at 21 frames per second. The comparison with state-of-the-art trackers (MEEM \cite{zhang2014meem}, HCF \cite{ma2015hierarchical}, LCT \cite{ma2015long}, fDSST \cite{danelljan2016discriminative}, Struck \cite{hare2016struck}, SRDCF \cite{danelljan2015learning}, Staple \cite{bertinetto2016staple}, KCF \cite{henriques2015high}) is shown in Figure \ref{comparison_TC}. Among the eight compared trackers, HCF \cite{ma2015hierarchical} obtains the best distance precision rate (DPR) of 70.5\% and Staple \cite{bertinetto2016staple} achieves the best overlap success rate (OSR) of 49.7\%. By comparison, PTAV improves the state-of-the-art methods on DPR to \textbf{74.1\%} and OSR to \textbf{54.4\%}, obtaining the gains of 3.6\% and 4.7\%, respectively.

Compared with fDSST~\cite{danelljan2016discriminative}, which obtains a DPR of 57.5\% and an OSR of 43.5\%, PTAV achieves significant improvements, showing clear the benefits of using a verifier. More results are left in the supplementary material.

\subsection{Experiment on UAV20L}
We also conduct experiment on the recently proposed UAV20L \cite{mueller2016benchmark} dataset that contains 20 videos. The shortest video contains 1,717 frames, and the longest one contains 5,527 frames. Our PTAV runs at 25 frames per second. The comparison with state-of-the-art trackers (MUSTer \cite{hong2015multi}, SRDCF \cite{danelljan2015learning}, HCF \cite{ma2015hierarchical}, MEEM \cite{zhang2014meem}, SAMF \cite{li2014scale}, Struck \cite{hare2016struck}, fDSST \cite{danelljan2016discriminative}, LCT \cite{ma2015long} and KCF \cite{henriques2015high}) is shown in Figure \ref{comparison_UAV20L}.  PTAV achieves the best performance on both the distance precision rate (62.4\%) and the success overlap rate (42.3\%), outperforming other approaches by large margins (11\% and 8\% compared with the second best in the distance precision rate and the success overlap rate, respectively).

\subsection{Detailed analysis of PTAV}
\noindent\textbf{Different verification interval $V$}.
In PTAV, different verification interval $V$ may affect both the accuracy and efficiency. A smaller $V$ means more frequent verification, which requires more computation and thus degrades the efficiency. A larger $V$, on the contrary, may cost less computation but may put PTAV at the risk when the target appearance change quickly. If the tracker loses the target object, it may update vast backgrounds in its appearance model until next verification. Even if the verifier re-locates the target and offers a correct detection result, the tracker may still lose it due to heavy changes of tracking appearance model. Table \ref{differentV} shows sampled results with different $V$ on OTB2013. Taking into account both accuracy and speed, we set $V$ to 10 in our experiments.

\vspace{.511mm}\noindent\textbf{Two threads v.s. one}.\footnote{Some ensemble tracking approaches can be implemented in multi-threads, \eg, TLD~\cite{kalal2012tracking}, MUSTer~\cite{hong2015multi} and LCT~\cite{ma2015long}). In such cases, different threads function similarly and almost independently. PTAV, by contrast, is fundamentally different. In PTAV, the tracking and verifying threads function very differently, and interact with each other frequently.}
In PTAV, the tracker does not rely on verifier in most time. To improve efficiency, we use two separate threads to process tracking and verifying in parallel instead of a single thread to process all tasks in a line. As a consequence, the tracker does not have to wait for the feedback from verifier to process next frame, and it traces back and resumes tracking only when receiving the positive feedback from verifier. Owing to storing all intermediate status, the tracker is able to quickly trace back without any extra computation. Table \ref{speed} shows the comparison of speed between using two threads and using only a single thread. From Table \ref{speed}, we can see that using two threads in parallel clearly improves the efficiency of system.

\renewcommand\arraystretch{1.3}
\begin{table}[!t]\footnotesize
  \centering
  \caption{Comparison of different $V$ in DPR and speed.}
    \begin{tabular}{rccc}
    \hline
          &  $V=5$ & $V=10$ & $V=15$ \\
    \hline
    DPR (\%) & 89.7    & 89.4    & 87.9 \\
    \hline
    Speed (\textit{fps}) & 23    & 27    & 29 \\
    \hline
    \end{tabular}%
  \label{differentV}%
\end{table}%

\renewcommand\arraystretch{1.3}
\begin{table}[!t]\footnotesize
  \centering
  \caption{Comparison on the tracking speed (\textit{fps}).}
    \begin{tabular}{@{}C{1.cm}@{} @{}C{1.85cm}@{}@{}C{1.8cm}@{}@{}C{1.5cm}@{}@{}C{1.7cm}@{}}
    \hline
    Threads      & OTB2013\cite{wu2013online} & OTB2015\cite{wu2015object} & TC128\cite{liang2015encoding} & UAV20L\cite{mueller2016benchmark} \\
    \hline
    One & 16    & 14    & 11 & 15\\
    \hline
    Two & 27    & 25    & 21 & 25\\
    \hline
    \end{tabular}%
  \label{speed}%
\end{table}%

\renewcommand\arraystretch{1.3}
\begin{table}[!t]\footnotesize
  \centering
  \caption{Comparison of different trackers in PTAV with $V=10$.}
    \begin{tabular}{c|c|cc}
    \hline
    \multicolumn{1}{r}{} & \multicolumn{1}{c}{} & PTAV with fDSST & PTAV with KCF \\
    \hline
    \multirow{3}[0]{*}{OTB2013} & DP (\%)   & 89.4  & 80.4 \\
          & OS (\%)   & 82.7  & 66.3 \\
    \cline{2-4}
          & Speed (\textit{fps})   & 27    & 24 \\
    \hline
    \multirow{3}[0]{*}{OTB2015} & DP (\%)   & 84.9  & 73.5 \\
          & OS (\%)   & 77.6  & 57.9 \\
    \cline{2-4}
          & Speed (\textit{fps})   & 25    & 21 \\
    \hline
    \end{tabular}%
  \label{comparison_OTB_diff_trackers}%
\end{table}%

\vspace{.511mm}\noindent\textbf{Different tracker $\TRK$}.
In PTAV, $\TRK$ is required to be efficient and accurate most of the time. To show the effects of different $\TRK$, we compare two different choices including fDSST~\cite{danelljan2016discriminative} and KCF~\cite{henriques2015high}. Compared to fDSST, KCF is more efficient while less accurate in short time interval. The comparison on OTB2013 and OTB2015 are provided in Table~\ref{comparison_OTB_diff_trackers}. It shows that PTAV with fDSST performs better and more efficiently than PTAV with KCF. Though KCF runs faster than fDSST, it performs less accurately in short time, which results in more requests for verifications and detections, hence significantly increasing the computation. By contrast, fDSST is more accurate in short time, and finally leads to efficiency in computation.

\section{Conclusion}

In this paper, we propose a new general tracking framework, named \emph{parallel tracking and verifying} (PTAV), which decomposes object tracking into two sub-tasks, tracking and verifying. We show that by carefully distributing the two tasks into two parallel threads and allowing them to work together, PTAV can achieve the best known tracking accuracy among all real-time tracking algorithms. The encouraging results are demonstrated in extensive experiments on four popular benchmarks. Since PTAV is a very flexible framework with great rooms for improvement and generalization, we expect this work to stimulate the designing of more efficient tracking algorithms in the future.

\small\vspace{1mm}\noindent\textbf{Acknowledgement}. This work is supported by National Key Research and Development Plan (Grant No. 2016YFB1001200) and National Nature Science Foundation of China (no. 61528204).

{\small
\bibliographystyle{ieee}
\bibliography{ptav}
}

\end{document}